\documentclass[12pt]{article}
\usepackage{graphicx}
\usepackage{amssymb}
\graphicspath{{figures/}}
\usepackage{float}
\usepackage{verbatim}
\usepackage[authoryear,round]{natbib}
\usepackage[ruled,vlined,linesnumbered]{algorithm2e}
\usepackage{geometry}
\usepackage{setspace}
\usepackage{authblk}
\restylefloat{figure}
\floatstyle{plaintop}
\restylefloat{table}
\usepackage[
  colorlinks=true,
  linkcolor=black,
  citecolor=black,
  urlcolor=blue,
  bookmarks=true,
  pdfauthor={Param Patel, Jay Dave, Pratyush Chakraborty},
  pdftitle={A Vendor-Agnostic LiDAR Data Conversion System}
]{hyperref}
\usepackage{booktabs}
\usepackage{url}
\usepackage{doi}

\title{\textbf{\large A Vendor-Agnostic LiDAR Data Conversion System with\\[2pt] Multi-Signal Detection and Multi-Format Output}}

\author[1]{Param Patel}
\author[2]{Jay Dave}
\author[1]{Pratyush Chakraborty\thanks{Corresponding author. Department of Electrical and Electronics Engineering, BITS Pilani, Hyderabad Campus, Telangana, India. Tel: +91 40 66303749. Email: pchakraborty@hyderabad.bits-pilani.ac.in}}

\affil[1]{Department of Electrical and Electronics Engineering, BITS Pilani, Hyderabad Campus, Telangana, India}
\affil[2]{Department of Computer Science and Information Systems, BITS Pilani, Hyderabad Campus, Telangana, India}

\date{}

\begin{document}

{\singlespacing
\maketitle
\vspace{0.5em}
\noindent\textbf{Highlights}
\begin{itemize}
\item Automatic detection of LiDAR vendor directly from raw PCAP files.
\item Multi-signal scoring across six packet-level features ensures reliable detection.
\item Vendor-specific wrappers convert raw data into five standard point cloud formats.
\item Pipeline evaluated on real outdoor captures from Ouster, Velodyne, Hesai, and Livox sensors across eight test files spanning both PCAP and vendor-proprietary formats.
\item C++ SDK-backed parsers achieve up to 10$\times$ higher throughput than Python-based parsers.
\end{itemize}
\vfill
}
\newpage

\begin{abstract}
LiDAR (Light Detection and Ranging) sensors capture the surrounding environment as dense 3D point clouds by measuring the time-of-flight of emitted laser pulses, making them foundational across autonomous vehicles, robotics, and large-scale mapping. PCAP (Packet Capture) files from these sensors are the starting point of most 3D perception pipelines, yet internal packet structures, UDP (User Datagram Protocol) port conventions, and encoding schemes differ enough across manufacturers that no single headless, zero-configuration tool reads them all without manual vendor selection. Ouster, Velodyne, Hesai, and Livox each require their own SDK (Software Development Kit), their own environment setup, and their own conversion workflow. Supporting all four means maintaining four disconnected pipelines with no shared infrastructure. The proposed pipeline accepts raw PCAP files as input and performs automated vendor identification by scoring six independent packet-level characteristics through a weighted multi-signal approach to determine the source sensor. C++ SDKs handle Ouster and Velodyne, while Hesai and Livox rely on Python-based dpkt parsing where no open source SDK exists. From there, a single command writes output to any of five industry-standard formats. We tested on real outdoor captures. Ouster peaks at 2.08M points per second, Velodyne at 1.47M, both running through native C++ packet decoding. Hesai and Livox land at 110K and 150K respectively, where Python-layer parsing introduces overhead that compounds under sustained load. The 8--10$\times$ gap held consistently across runs. Benchmarks were conducted on a consumer-grade Intel i3 workstation with 8~GB RAM running Windows~11, deliberately chosen to demonstrate that the pipeline operates without server-class hardware or a Linux-only environment --- a practical consideration for field deployment scenarios where dedicated processing infrastructure may not be available.
\end{abstract}

\noindent\textbf{Keywords:} LiDAR; PCAP; point cloud; vendor detection; data conversion; Ouster; Velodyne; Hesai; Livox

\clearpage
\section{Introduction}
\label{introduction}

LiDAR (Light Detection and Ranging) serves as a foundational sensing technology for robotics, smart infrastructure, and autonomous navigation \citep{Roriz2022}. Unlike passive camera-based systems, LiDAR maintains high spatial accuracy regardless of ambient lighting conditions or common environmental occlusions. By measuring the time-of-flight of emitted laser pulses to reconstruct the surrounding geometry, the sensor provides a high-fidelity 3D representation that is functionally more robust than vision centric alternatives in dynamic or low-visibility settings. What comes out of these sensors is raw 3D point cloud data, and the sensor itself stores it directly in a format called PCAP \citep{Harris2022}. There are very few sensor manufacturing companies in this domain, and among them, Ouster, Velodyne, Hesai, and Livox are the primary players \citep{Roriz2022}. All these companies produce different kinds of sensors that give different characteristics and data densities of the point cloud data in PCAP format. This raw point cloud data needs to be converted into standard formats to be used in a machine learning pipeline, a mapping system, or a robotics framework.

Heterogeneity across vendors compounds this further. Each stores raw data with different UDP structures \citep{Lin2017}, port conventions, and encoding schemes, so a tool built for one vendor cannot be reused for another. SDK dependencies add a second layer of friction: vendor toolchains require independent runtime environments and hardware assumptions, and Hesai and Livox provide no production-grade open-source Python SDKs at all, forcing practitioners to maintain separate, fragile workflows per manufacturer.

Several tools exist for point cloud processing and LiDAR visualisation. PDAL \citep{Butler2021} and ROS-based tools \citep{Quigley2009} handle already-decoded formats well but have no raw PCAP ingestion capability. Vendor-provided viewers like Ouster Studio, VeloView, and Livox Viewer handle visualisation within their own ecosystems, while LidarView \citep{Kitware2025LidarView} supports all four vendors but requires manual vendor or calibration selection per file. None provide a headless, zero-configuration path from raw PCAP to standard file formats.

Vendor-provided SDKs offer one path forward, but each requires independent environment configuration, and the operator must first identify which vendor produced the data before any SDK can be selected. This identification and setup process must be repeated independently for each manufacturer encountered in a heterogeneous deployment. State-of-the-art tools that do support multiple vendors, such as LidarView, still require the operator to manually select the appropriate calibration or interpreter before processing begins \citep{Roriz2022}. Automated, zero-configuration vendor detection from raw packet content remains an unaddressed gap.

The proposed system addresses this gap directly: given a raw PCAP file, it performs vendor identification and format conversion autonomously. It selects the appropriate parser --- a C++ SDK for Ouster \citep{Ouster2024SDK} and Velodyne \citep{Valgur2024}, and a purely Python-based one using dpkt \citep{Song2024dpkt} for Hesai and Livox --- to convert the raw PCAP file into five industry-accepted formats from a single command. The whole system works on consumer-grade hardware and requires no prior knowledge of the vendor or format, making it widely accessible. The complete source code is publicly available at \citep{PatelGitHub2024}.

The remainder of this paper is organised as follows: Section~\ref{problem_statement} defines the problem and goals, Section~\ref{literature_review} reviews related work, Section~\ref{proposed_approach} presents the proposed approach, Section~\ref{performance_analysis} covers the experimental analysis, Section 6 mention threats to validity and Section~\ref{conclusion} concludes with future directions.

\section{Problem Statement}
\label{problem_statement}

\subsection{System Model}

The system consists of two main parts.

The User is anyone working with raw LiDAR data, usually in .pcap \citep{Harris2022} or other vendor-specific formats, who needs to convert it into usable point cloud formats. This is common in areas like robotics, computer vision, 3D mapping, and autonomous systems \citep{Roriz2022}, where raw sensor data is not directly usable.

The Conversion System is a pipeline that takes this raw data as input, detects the sensor vendor automatically, and converts it into standard formats like .pcd \citep{Rusu2024PCD} (Point Cloud Data), .csv (Comma-Separated Values), .las (LiDAR Aerial Survey) \citep{ASPRS2019}, and others.

\begin{figure}[H]
\centering
\includegraphics[width=\linewidth]{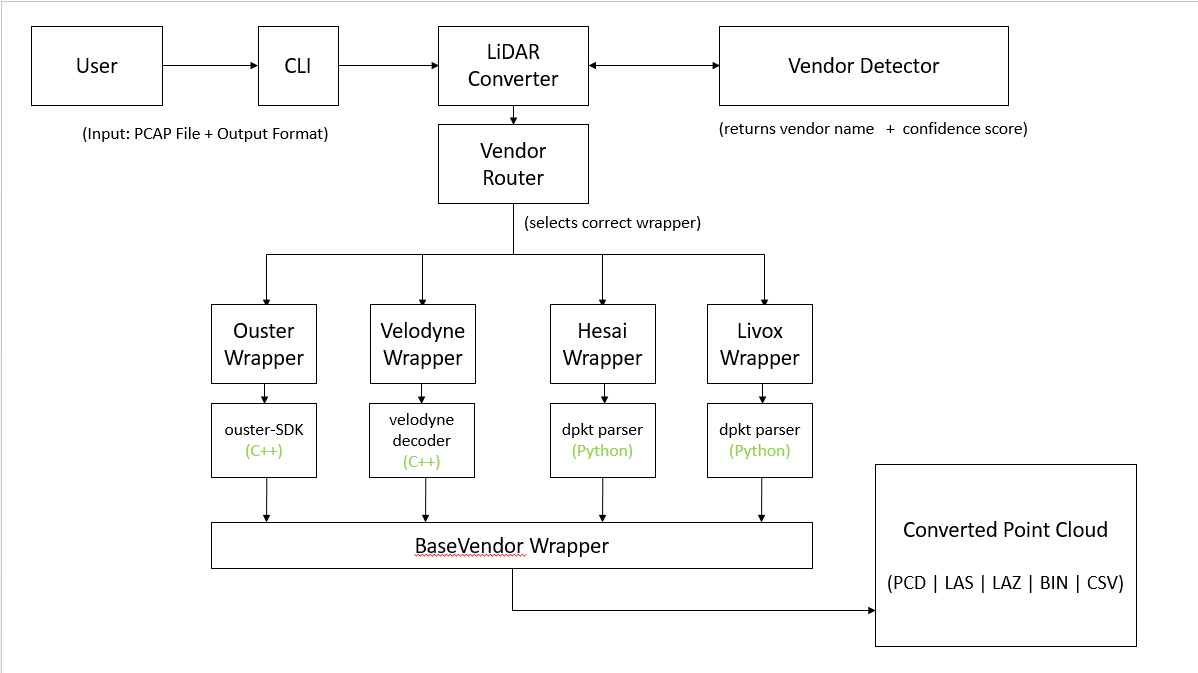}
\caption{System architecture overview, showing the flow from user input through the Vendor Detector, vendor-specific wrappers, and the Base Wrapper to produce converted point cloud output.}
\label{fig:architecture}
\end{figure}

Figure~\ref{fig:architecture} illustrates the overall component layout of the system, showing the flow from the user's input through vendor detection, the vendor-specific wrappers, and the Base Wrapper to produce the converted point cloud output. The internal layering of this architecture is discussed in detail in Section~\ref{proposed_approach}.

\subsection{Problem Definitions}

\textit{Packet-Level Heterogeneity:} LiDAR workflows are currently fragmented at the transport layer. While Ouster, Velodyne, Hesai, and Livox all utilize PCAP containers \citep{Harris2022}, their internal UDP layouts \citep{Lin2017}, firing sequences, and port conventions differ enough that a parser built for one vendor cannot be reused for another without significant modification. Without a shared industry standard, perception stacks using multiple sensors must maintain largely isolated parsing logic, creating a maintenance burden where little code is shared between manufacturer-specific modules.

\textit{Toolchain and SDK Dependency:} Interfacing with raw data typically requires vendor-specific SDKs \citep{Ouster2024SDK,Valgur2024}, each introducing unique runtime dependencies and OS-specific assumptions. This fragmentation is most acute when official support is incomplete; for instance, Hesai and Livox lack stable, open-source Python SDKs for direct conversion. We observed that this forces engineers to choose between writing fragile custom parsers or relying on third-party tools that fail when sensor firmware or packet lengths change.

\textit{Cross-Domain Output Friction:} Robotics \citep{Quigley2009}, machine learning, and geomatics ecosystems do not share a common data currency. A single raw PCAP might need to be exported as a KITTI-style binary \citep{Geiger2012} for training and simultaneously as a LAS file \citep{ASPRS2019} for geographic archival \citep{LoC2024LAS}. Navigating these transitions usually involves a ``tool-chain hop,'' adding computational overhead and increasing the risk of coordinate transform errors.

\subsection{Goals}

\begin{itemize}
\item \textbf{Zero-Configuration Detection:} Identify the hardware vendor directly from the raw bitstream \citep{Harris2022} to eliminate manual flags or user-defined parameters.

\item \textbf{Interface Consolidation:} Provide a single execution path to export PCD \citep{Rusu2024PCD}, LAS \citep{ASPRS2019}, LAZ \citep{LASzip2025}, BIN, and CSV, removing the need for auxiliary converters.

\item \textbf{Optimized Parsing Paths:} Maximize throughput by leveraging native C++ SDKs (Ouster \citep{Ouster2024SDK}, Velodyne \citep{Valgur2024}) where possible, while implementing lightweight dpkt \citep{Song2024dpkt} Python logic for unsupported vendors.
\end{itemize}

\section{Literature Review}
\label{literature_review}

\subsection{LiDAR Data Challenges in Multi-Vendor Environments}
\label{sec:multivendor}

Abbasi et al. survey the computational overhead of 3D point cloud processing for autonomous navigation, identifying that fluctuating point densities, beam counts, and noise profiles across LiDAR vendors create a significant bottleneck for unified perception stacks \citep{Abbasi2022}. The proposed pipeline addresses this vendor heterogeneity at the ingestion stage, standardising raw PCAP streams before they reach any downstream processing or learning layer.

Chen et al. captured simultaneous point clouds from Ouster 128, Ouster 64, and Hesai XT32 sensors across campus and urban scenarios, releasing the MLDAS dataset \citep{Chen2024}. The core problem they target is that models trained on one LiDAR generalise poorly to another due to structural differences in beam count, field of view, and detection range. Their solution, HSSC, is a hierarchical network using spatial-temporal consistency to transfer knowledge across sensor types \citep{Chen2024}. The work confirms that Ouster and Hesai, two of the four vendors supported by the proposed pipeline, differ significantly at the sensor level. Where Chen et al. address multi-vendor variability at the model level, the proposed system handles it earlier, at the raw data ingestion stage.

\subsection{Point Cloud Format and Storage Considerations}
\label{sec:formats}

Output format selection in the proposed pipeline is informed by two relevant surveys. Roriz et al. survey compression techniques for automotive LiDAR point clouds across lossless, lossy, static, and dynamic methods, observing that LAZ \citep{LASzip2025} offers a strong balance between lossless compression and file size reduction, BIN provides faster raw data access, and PCD \citep{Rusu2024PCD} is more flexible for storing multiple attributes \citep{Roriz2024}. These properties directly informed the five output formats supported by the proposed pipeline. Ladra et al. propose a data structure that simultaneously compresses and indexes spatial dimensions and attributes, enabling range and attribute queries directly on compressed data, up to 100 times faster than LAZ \citep{Ladra2024}. Both works operate on already-decoded data and do not address PCAP ingestion or vendor detection; the raw decoding stage preceding their inputs remains unaddressed in the toolchain.

\subsection{Point Cloud Processing Tools and Libraries}
\label{sec:processing_tools}

PDAL is a format-agnostic C++ framework for translating and manipulating point cloud data \citep{Butler2021}. Its core design chains independent processing stages such as reading, spatial filtering, and coordinate re-projection into modular pipelines executable from the command line without additional infrastructure. It handles LAS \citep{ASPRS2019}, LAZ \citep{LASzip2025}, PCD \citep{Rusu2024PCD}, PLY \citep{LoC2024PLY}, CSV, and BPF \citep{PDAL2024BPF}, formats that overlap with the outputs of the proposed system. However, PDAL operates on already-decoded point cloud files and provides no mechanism for raw PCAP ingestion, vendor identification, or packet-level decoding. The proposed system fills this upstream gap, producing the decoded standard-format files that PDAL and similar tools then consume.

Wang et al. proposed an end-to-end point cloud processing system that introduces a new data format, .PcRecord, designed to reduce storage size and improve loading efficiency \citep{Wang2025}. The system standardises input formats such as PCD \citep{Rusu2024PCD} or LAS \citep{ASPRS2019} into .PcRecord through a three-module architecture: a format normalisation module, a multi-stage parallel processing pipeline, and an autotune module that adjusts parameters based on real-time hardware metrics. This approach yields significant speedups of 6.61$\times$ on ModelNet40, 3.09$\times$ on KITTI \citep{Geiger2012}, and up to 25.4$\times$ on certain datasets \citep{Wang2025}. Like PDAL, however, this system assumes input data has already been decoded into a standard format; raw PCAP ingestion and vendor detection fall outside its scope.

\subsection{Existing LiDAR Viewer and Conversion Tools}
\label{sec:viewers}

Several tools exist that address LiDAR data visualisation and, to varying degrees, format conversion. LidarView, developed by Kitware, supports live streaming and offline playback of PCAP files from Ouster, Velodyne, Hesai, and Livox sensors, and provides an interactive 3D viewer with time-scrubbing and per-frame inspection \citep{Kitware2025LidarView}. Vendor-specific desktop tools offer similar capabilities within their own ecosystems: Ouster Studio supports Ouster sensors with calibration-aware visualisation \citep{Ouster2024SDK}, VeloView handles Velodyne PCAP playback \citep{Kitware2024VeloView}, and the Livox Viewer supports LVX and PCAP captures from Livox hardware \citep{Livox2023HAP}. These tools have meaningfully lowered the barrier to inspecting raw LiDAR data and established PCAP as a practical archival format across the industry.

These tools are, however, designed around interactive GUI workflows. LidarView requires the user to manually select the correct interpreter or calibration plugin before a file can be opened, meaning the vendor must already be known. None of them expose a command-line interface for batch or automated use. On the output side, LidarView and VeloView do not produce BIN format files, which is the binary format used by the KITTI benchmark \citep{Geiger2012} and widely expected by object detection pipelines. LidarView's export options are limited to CSV and a small set of proprietary formats, while PCD export requires a separate plugin and manual per-session steps. Ouster Studio and VeloView do not support cross-vendor files at all. In practice, when PCAP files arrive from mixed sensor deployments without per-file metadata identifying the source sensor, none of these tools can process them without operator intervention at every step.

The proposed system takes a different approach to this workflow. Rather than requiring the operator to identify the vendor upfront, it derives the vendor automatically from the packet stream itself using a multi-signal scoring mechanism across six file-level features. Where viewer tools render the point cloud visually for human inspection, the proposed system writes the decoded data directly to BIN, PCD, LAS, LAZ, and PLY files, making it a practical ingestion step for processing frameworks and ML pipelines that require standard format inputs. It is not a replacement for LidarView or Ouster Studio for interactive inspection; it is better understood as a complementary tool that serves the file-output side of the same raw PCAP input.

Table~\ref{tab:tool_comparison} summarises the key capability differences across tools in this space.

\begin{table}[H]
\centering

\label{tab:tool_comparison}
\begin{tabular}{lcccccc}
\hline
\textbf{Feature} & \textbf{Proposed} & \textbf{LidarView} & \textbf{PDAL} & \textbf{VeloView} & \textbf{Ouster Studio}  \\
\hline
Raw PCAP input     & \checkmark & \checkmark & $\times$   & \checkmark & \checkmark  \\
Auto vendor detect & \checkmark & $\times$   & $\times$   & $\times$   & $\times$     \\
Headless CLI       & \checkmark & $\times$   & \checkmark & $\times$   & $\times$     \\
Multi-vendor       & \checkmark & \checkmark & \checkmark & $\times$   & $\times$      \\
BIN/KITTI output   & \checkmark & $\times$   & $\times$   & $\times$   & $\times$     \\
LAS/LAZ output     & \checkmark & $\times$   & \checkmark & $\times$   & $\times$     \\
PCD output         & \checkmark & $\times$   & \checkmark & $\times$   & $\times$     \\
Batch/no GUI       & \checkmark & $\times$   & \checkmark & $\times$   & $\times$      \\
\hline
\end{tabular}
\end{table}

\section{Proposed Approach}
\label{proposed_approach}

Raw PCAP ingestion \citep{Harris2022} drives the design --- vendor identification, parser selection, and format conversion all follow from that single entry point. Automatic detection handles vendor identification through a multi-signal scoring mechanism, after which the data stream is routed to the appropriate parser backend.

\subsection{System Architecture Overview}

The system architecture follows a three-layer split: the Detector Layer, the Conversion Layer, and the Orchestration Layer. While the user interacts through a unified CLI, internal processing is segmented to allow modular vendor support.

The Orchestration Layer serves as the system's entry point. It manages execution state and passes input parameters such as max\_scans to downstream parsers.

The Detector Layer analyzes raw packet headers and payloads using a module called the Vendor Detector. It does not rely on a single flag and instead computes confidence scores across six independent file-level signals to reduce misidentification in multi-sensor captures.

The Conversion Layer handles format translation through vendor-specific wrappers, the Base Wrapper, and the Converter module. All these modules work internally on the raw data, while the Detector Layer acts as an input to this layer. Figure~\ref{fig:architecture} (Section~\ref{problem_statement}) illustrates this component layout.

\subsection{Vendor Detection Module}
\label{sec:vendor_detection}

This module detects the vendor from a raw PCAP file \citep{Harris2022} by deriving six signals from it and assigning scores to different vendors. The vendor with the maximum score that crosses a defined threshold is selected as the detected vendor. All these signals have different values for each vendor, and based on those values, the module determines the vendor. Thus, having the same final score for two vendors after evaluation is not possible.

\begin{algorithm}[H]
\caption{DetectVendor(F)}
\label{alg:detectvendor}
\KwIn{File path $F$}
\KwOut{vendor\_name $v$, confidence $c$, or NULL}
\If{$F$ does not exist or $F$ is empty}{
    \Return NULL\;
}
\If{extension($F$) = ``.rxp''}{
    \Return (``riegl'', confidence=1.0)\;
}
$scores \leftarrow \{v : 0.0 \mid v \in \mathit{VendorRegistry}\}$\;
$E \leftarrow$ CheckExtension($F$)\;
$scores \leftarrow scores + E \times 0.5$\;
$M \leftarrow$ CheckMagicBytes($F$)\;
$scores \leftarrow scores + M \times 3.0$\;
$C \leftarrow$ CheckCompanionFiles($F$)\;
$scores \leftarrow scores + C \times 2.5$\;
\If{extension($F$) = ``.pcap''}{
    $P \leftarrow$ CheckUDPPorts($F$)\;
    $scores \leftarrow scores + P \times 3.5$\;
    $S \leftarrow$ CheckPacketStructure($F$)\;
    $scores \leftarrow scores + S \times 3.0$\;
    $Z \leftarrow$ CheckPacketSize($F$)\;
    $scores \leftarrow scores + Z \times 2.0$\;
}
$v^* \leftarrow \mathrm{argmax}(scores)$\;
$raw\_score \leftarrow scores[v^*]$\;
\If{$raw\_score < 2.0$}{
    \Return NULL\;
}
$c \leftarrow \min(raw\_score / 14.5,\ 1.0)$\;
\Return $(v^*, c)$\;
\end{algorithm}

The six signals used to assign a score to each vendor are listed below with their respective weights. Not all signals are equally unique; some are common across multiple vendors, while others have discrete values specific to a single vendor. Signals with more vendor-specific and discrete values are given higher weights compared to those that are shared or ambiguous. The weight values were determined empirically through an iterative process: starting with equal weights, each signal was evaluated in isolation across the eight test files, and weights were incrementally adjusted based on each signal's vendor-discriminative power. The process continued until detection remained correct across all test cases, including the Velodyne--Hesai port~2368 overlap scenario where UDP port alone is insufficient. No formal optimisation was applied; the weights represent a manually validated configuration over the available test set.

\textbf{File Extension (weight: 0.5)} --- We check for .pcap \citep{Harris2022}, .lvx \citep{Livox2019LVX}, or .rxp formats. Filenames often contain strings like ``velodyne'' or ``ouster'' which provide initial hints. This receives the lowest weight because extensions are easily modified and don't guarantee the internal data structure.

\textbf{Magic Bytes (weight: 3.0)} --- The system reads the first 1,024 bytes to find manufacturer signatures. We match against 0xFFEE for Velodyne \citep{Valgur2024}, 0x0001 for Ouster \citep{Ouster2024SDK}, 0xEEFF for Hesai \citep{Hesai2025}, and the ``livox\_tech'' ASCII string for Livox \citep{Livox2019LVX}.

\textbf{Companion Files (weight: 2.5)} --- This check scans for sidecar metadata in the same directory. Ouster sensors, for instance, require a specific .json file \citep{Ouster2024SDK} to interpret the raw stream \citep{Ouster2024Python}. Validating these files provides a strong vendor indicator that raw packets alone might miss.

\textbf{UDP Port Detection (weight: 3.5)} --- Highest weight. Parses the first 10 UDP packets \citep{Song2024dpkt} and checks the destination port. Velodyne \citep{Valgur2024}: 2368/2369 \citep{Rusu2024HDL}, Ouster \citep{Ouster2024SDK}: 7502/7503 \citep{Ouster2024SensorData}, Hesai \citep{Hesai2025}: 2368, Livox \citep{Livox2019LVX}: 56000--58000 \citep{Livox2023HAP}.

\textbf{Packet Structure (weight: 3.0)} --- Reads first 2 bytes of UDP payload per packet and matches against vendor magic bytes \citep{Song2024dpkt}. More reliable than file-level magic bytes since it confirms packet-level structure.

\textbf{Packet Size (weight: 2.0)} --- Samples first 50 UDP payloads \citep{Song2024dpkt}, checks if 80\% or more fall within vendor-specific size ranges. Velodyne \citep{Valgur2024}: exactly 1206 bytes, Ouster \citep{Ouster2024SDK}: 6400--33024 bytes, Hesai \citep{Hesai2025}: 1000--1300 bytes. While Velodyne's typical packet size of 1206 bytes falls within Hesai's expected range of 1000--1300 bytes, making packet size alone insufficient to distinguish this pair, the signal contributes meaningfully when combined with the other five signals, particularly UDP port and magic byte checks, which are unambiguous for this vendor pair.

\begin{algorithm}[H]
\caption{CheckUDPPorts(F)}
\label{alg:checkudpports}
\KwIn{PCAP file path $F$}
\KwOut{scores dictionary per vendor}
$scores \leftarrow \{\}$\;
$ports\_found \leftarrow \{\}$\;
$packet\_count \leftarrow 0$\;
\ForEach{$(timestamp, buffer)$ in pcap.Reader($F$)}{
    \If{$packet\_count > 10$}{
        \textbf{break}\;
    }
    parse Ethernet $\rightarrow$ IP $\rightarrow$ UDP from $buffer$\;
    $dst\_port \leftarrow$ UDP.destination\_port\;
    \ForEach{vendor $v$ in VendorRegistry}{
        \If{$dst\_port \in v.\mathit{udp\_ports}$}{
            $ports\_found[v]$.append($dst\_port$)\;
        }
    }
    $packet\_count \leftarrow packet\_count + 1$\;
}
\ForEach{$v$ with $ports\_found[v] \neq \emptyset$}{
    $scores[v] \leftarrow 1.0$\;
}
\Return $scores$\;
\end{algorithm}

Among the six signals used to deduce the vendor of the data, UDP port detection is given the highest weight because it is the most vendor-distinct signal across the four vendors \citep{Rusu2024HDL,Ouster2024SensorData,Livox2023HAP,Hesai2025}. Although Velodyne and Hesai share port 2368, this overlap is resolved in practice by the combined contribution of the remaining signals, particularly packet structure and magic bytes, which are unambiguous for this vendor pair. No two vendors produce an identical score across all six signals simultaneously, making the final detection deterministic.

\subsection{Conversion Layer}

\subsubsection{Converter Module}

This module coordinates between the Detector Layer and the appropriate parsing backend for the identified vendor. It does not convert data by itself.

The module receives the input file \citep{Harris2022} along with the desired output format. It runs basic validation first --- file existence checks, corruption handling, and format support verification.

After validation, the Vendor Detector module is invoked and the detected vendor is used to select the corresponding parsing wrapper. The module then confirms whether a compatible parsing backend exists and whether the required SDKs \citep{Ouster2024SDK,Valgur2024} are available for execution.

Format support is checked separately, since PCD \citep{Rusu2024PCD}, LAS \citep{ASPRS2019}, LAZ \citep{LASzip2025}, BIN \citep{Geiger2012}, and CSV availability varies across vendor SDKs. Only after all conditions are satisfied is the file passed to the corresponding vendor-specific wrapper for conversion, detailed in the next section.

The output includes a summary of execution: processing time, number of converted points, vendor confidence score, and any runtime warnings or errors.

\begin{algorithm}[H]
\caption{Convert(F, format)}
\label{alg:convert}
\KwIn{File path $F$, output format}
\KwOut{Result dictionary $R$}
Validate $F$ exists, non-empty, format is supported\;
\If{validation fails}{
    \Return error result\;
}
Generate output path from $F$ and format\;
$detection\_result \leftarrow$ DetectVendor($F$)\;
\If{$detection\_result$.success = FALSE}{
    \Return error result\;
}
$vendor \leftarrow detection\_result$.vendor\_name\;
$wrapper \leftarrow$ GetWrapper($vendor$)\;
\If{$wrapper$ = NULL}{
    \Return error result\;
}
\If{$wrapper$.sdk\_available = FALSE}{
    \Return error result\;
}
\If{$format \notin wrapper$.supported\_formats}{
    \Return error result\;
}
\If{$format$ = ``las''}{
    $conversion\_result \leftarrow wrapper$.ConvertToLAS($F$)\;
}
\Else{
    $conversion\_result \leftarrow wrapper$.Convert($F$, $format$)\;
}
\If{$conversion\_result$.success = TRUE}{
    validate output file\;
    \Return success result with points\_converted\;
}
\Else{
    \Return error result\;
}
\end{algorithm}

\subsubsection{Base Vendor Module}

This abstract class establishes a standardized template for every vendor-specific implementation. It enforces six mandatory methods --- get\_vendor\_name, validate\_sdk\_installation, convert\_to\_las, convert, get\_vendor\_info, and validate\_conversion --- shared across wrappers to avoid duplication. This interface decouples the Converter module from the specific parsing logic of different manufacturers. By interacting only through these base methods, the pipeline triggers diverse conversion paths for Ouster \citep{Ouster2024SDK} or Livox \citep{Livox2019LVX} through a single execution flow, regardless of the underlying SDK requirements. It also includes three shared utility functions for serialisation --- pointsto\_pcd \citep{Rusu2024PCD}, pointsto\_bin \citep{Geiger2012}, and pointsto\_csv --- which are reused across wrappers to reduce repeated code. LAZ \citep{LASzip2025} compression is handled in this module using a fallback chain. It first tries the built-in laspy \citep{laspy2024} compression, and if that fails, it switches to an external laszip \citep{LASzip2025} binary. If both fail, the system does not stop the conversion; instead, it outputs the LAS \citep{ASPRS2019} file and raises a warning. This ensures that an output file is always generated, even when compression fails.

\subsubsection{Vendor-Specific Wrappers}

All four wrappers inherit the same methods from the Base Wrapper and follow the same interface. The main difference is in how each one handles parsing internally. Some use C++-based SDKs, while others rely on Python-based parsing for raw data. This mainly comes from the fact that Hesai and Livox do not provide open-source Python SDK support for conversion. Even with these differences, the output from all four wrappers is the same: an N$\times$4 float32 NumPy \citep{Harris2020NumPy} array carrying x, y, z, and intensity values per point. Additional per-point fields available from vendor SDKs, such as ring index and per-point timestamps, are not currently extracted; extending the output schema to include these fields is a straightforward addition identified for future work. This array is then passed to the Base Wrapper utility functions for serialisation.

\subsubsection*{(1) Ouster Wrapper}

Ouster produces the highest throughput among the four vendors. A major reason for this is the Ouster SDK \citep{Ouster2024SDK}, which is C++ based with Python bindings and integrates well with the pipeline. Unlike other vendors, the ouster-sdk \citep{Ouster2024SDK} requires a JSON metadata file along with the raw PCAP file \citep{Harris2022}. This file contains beam angles and sensor calibration settings \citep{Ouster2024Python}, which the SDK uses to build an XYZ lookup table. This removes the need for per-point trigonometric calculations and improves parsing speed. The SDK uses OusterPacketSource and ScanBatcher for packet reading and scan assembly, both handled internally in C++. It also handles destaggering internally, unlike the Python-based implementations where it is done manually. Scans are accumulated and stacked into an N$\times$4 float32 NumPy \citep{Harris2020NumPy} array, which is then passed to the Base Wrapper utility functions.

\subsubsection*{(2) Velodyne Wrapper}

Velodyne uses a C++ backed library called velodyne-decoder \citep{Valgur2024} for converting raw PCAP data \citep{Harris2022}, but unlike Ouster, it does not require a JSON metadata file. Because of this, the SDK reads the PCAP file through an iterator that returns one scan at a time as an N$\times$8 float32 NumPy \citep{Harris2020NumPy} array, and the wrapper keeps only the x, y, z, and intensity columns. Coordinate conversion still relies on per-model elevation angle lookup tables stored in the wrapper. Unlike Ouster's XYZLut approach, trigonometry is computed per point instead of being precomputed, which contributes to lower throughput. The wrapper supports multiple sensor models, including VLP-16, VLP-32C, HDL-32E, HDL-64E, and VLS-128 \citep{Velodyne2019}, each with different channel counts and elevation angles. The correct model is auto-detected before conversion by inspecting magic bytes and packet structure \citep{Valgur2024}. If velodyne-decoder is not available, a Python-based dpkt \citep{Song2024dpkt} fallback is used, so the pipeline still works for Velodyne data even without the SDK.

\subsubsection*{(3) Hesai Wrapper}

Since there is no open-source C++ SDK available for Hesai, a purely Python-based dpkt \citep{Song2024dpkt} approach is used to parse its proprietary packet structures \citep{Hesai2025}. Each packet is identified by checking the header, where the signature 0xEEFF is used to mark Hesai packets \citep{Hesai2025}. To compute x, y, z, and intensity, per-channel elevation and azimuth tables are hardcoded inside the wrapper, based on the vendor's sensor documentation \citep{Hesai2025}. The wrapper supports most of the Pandar series LiDAR sensors from Hesai \citep{Hesai2025}. This Python-based implementation is also the reason for lower throughput compared to Ouster and Velodyne, which use C++ SDKs.

\subsubsection*{(4) Livox Wrapper}

Unlike the other three vendors, Livox is different because it supports three types of inputs --- raw PCAP files \citep{Harris2022}, CSV files exported from Livox Viewer \citep{Livox2020Viewer}, and proprietary formats called LVX and LVX2 \citep{Livox2019LVX}. For PCAP files, the parsing approach is similar to Hesai, using a purely Python-based dpkt \citep{Song2024dpkt} implementation, so throughput is also similar and lower than Ouster and Velodyne. Output data follows the N$\times$4 float32 NumPy \citep{Harris2020NumPy} array format required by the Base Wrapper. The main bottleneck here is the proprietary LVX/LVX2 binary structure \citep{Livox2019LVX}. Since there is no public C++ SDK, we had to build a bit-level parser from scratch using Python's struct module \citep{Python2024struct}. We manually map the file header, device info, and frame data blocks. This manual approach makes Livox the most complex vendor to support architecturally, and the bit-manipulation overhead in Python is exactly why the throughput stays low compared to C++-based vendors.

\section{Performance Analysis}
\label{performance_analysis}

Pipeline performance was evaluated across datasets of different sizes on a single workstation under varying workloads. The setup is kept fixed so the results are consistent across runs. The evaluation focuses on whether a unified pipeline can efficiently support vendors whose SDK architectures and parsing workflows differ substantially across Python- and C++-based implementations. For Ouster and Velodyne, the vendor wrappers expose the native C++ SDK memory buffers directly to Python via the SDK's own optimised bindings, introducing no intermediate serialisation or buffering layer. The observed peak throughput of 2.08M points/sec (Ouster) and 1.47M points/sec (Velodyne) reflects the physical decoding limits of the bare SDKs on the evaluation hardware, confirming that the unified abstraction layer introduces no measurable computational overhead for C++-backed vendors. For Hesai and Livox, throughput is bounded by Python-layer packet parsing, which is an SDK availability constraint rather than a pipeline design choice.

\subsection{Experimental Setup}
\label{sec:experimental_setup}

\subsubsection{Hardware Specs of the Workstation}

The pipeline is executed on a system with an Intel Core i3-1005G1 processor \citep{Intel2024} with a base clock speed of 1.20 GHz and a boost clock speed of 3.40 GHz. The system contains 8 GB DDR4 RAM at 3200 MHz. There is no involvement of a GPU; the whole pipeline executes as a CPU-bound process. The primary storage used is a 256 GB SSD, and the operating system used to implement the pipeline is Windows 11. The system is deliberately kept as modest consumer-grade hardware to demonstrate that the pipeline does not require specialised infrastructure.

\subsubsection{Software Environment}

The pipeline is built with Python 3.13.3 \citep{Python2024Docs}, commonly used in robotics \citep{Quigley2009} and point cloud processing workflows in autonomous vehicles \citep{Roriz2022}, which makes it easier to integrate with the existing workflow systems of these industries. For point cloud parsing and conversion, we use primarily two types of technologies: the SDKs that are C++ based with Python bindings provided by vendors, namely ouster-sdk 0.15.1 \citep{Ouster2024SDK} and velodyne-decoder 3.1.0 \citep{Valgur2024}, and purely Python-based dpkt 1.9.7 \citep{Song2024dpkt}, a packet parsing library used for vendors like Hesai and Livox. Additionally, libraries like laspy 2.6.0 \citep{laspy2024} for conversion and extraction of data in formats like LAS \citep{ASPRS2019} and LAZ \citep{LASzip2025} across all four vendors, and NumPy \citep{Harris2020NumPy}, handle all the memory and data extraction related functions. Additional supporting libraries include pandas \citep{McKinney2010} for Livox CSV ingestion, and rich and tqdm for CLI progress reporting.

\subsubsection{Point Cloud Datasets}

Sourcing raw, uncompressed PCAP files directly from physical LiDAR sensors presents significant practical challenges: most public repositories distribute data in ROS bag format, vendor proprietary formats, or already-decoded point cloud files rather than raw packet captures, and Hesai in particular has very limited open PCAP availability. To ensure a rigorous evaluation despite these constraints, we curated a diverse benchmark set of eight files spanning both standard PCAP and vendor-proprietary formats. The set deliberately covers a wide operational spectrum: channel counts ranging from 16 (VLP-16) to 128 (VLS-128, OS-2-128), distinct packet structures across all four manufacturers, one proprietary binary format (Livox LVX2), and varied high-noise outdoor environments including busy road intersections, drone flights, and urban drives. This structural diversity serves as a comprehensive stress-test for the multi-signal detection algorithm and the vendor-specific parsing backends.

Files are used across two evaluation categories: detection evaluation (all eight files) and throughput benchmarking (one file per vendor). Table~\ref{tab:detection_files} lists the detection set and Table~\ref{tab:throughput_files} lists the throughput set.

\begin{table}[H]
\centering
\caption{Files used for vendor detection evaluation.}
\label{tab:detection_files}
\begin{tabular}{clllr}
\hline
\textbf{\#} & \textbf{Filename} & \textbf{Vendor} & \textbf{Format} & \textbf{Size} \\
\hline
1 & hesai\_BusyRoad.pcap              & Hesai    & .pcap & 3.48 GB \\
2 & Outdoor\_sampledata.lvx2          & Livox    & .lvx2 & 569 MB  \\
3 & StaticCarIntersection\_Livox-HAP.pcap & Livox & .pcap & 604 MB  \\
4 & OS-2-128\_..\_100425-000.pcap & Ouster & .pcap & 4.28 GB \\
5 & Urban\_Drive\_.pcap               & Ouster   & .pcap & 4.80 GB \\
6 & CarLoop\_Velodyne-VLP16.pcap      & Velodyne & .pcap & 134 MB  \\
7 & Drone\_Velodyne-HDL32.pcap        & Velodyne & .pcap & 649 MB  \\
8 & OfficeWalking\_Velodyne-VLS128.pcap & Velodyne & .pcap & 2.32 GB \\
\hline
\end{tabular}
\end{table}

\begin{table}[H]
\centering
\caption{Files used for throughput benchmarking, one per vendor. Ouster requires a companion JSON metadata file for SDK calibration.}
\label{tab:throughput_files}
\begin{tabular}{lllr}
\hline
\textbf{Filename} & \textbf{Vendor} & \textbf{Sensor Model} & \textbf{Size} \\
\hline
Urban\_Drive\_.pcap                    & Ouster   & OS-2-128    & 4.80 GB \\
OfficeWalking\_Velodyne-VLS128.pcap    & Velodyne & VLS-128     & 2.32 GB \\
hesai\_BusyRoad.pcap                  & Hesai    & Pandar      & 3.48 GB \\
StaticCarIntersection\_Livox-HAP.pcap & Livox    & HAP         & 604 MB  \\
\hline
\end{tabular}
\end{table}

All files are outdoor capture scenarios, deliberately chosen to test the pipeline under real-world operating conditions where environmental noise, dynamic objects, and varying point densities are naturally present, factors that stress-test the pipeline beyond controlled indoor environments \citep{Roriz2022}.

\subsection{Evaluation Metrics}

This pipeline is evaluated on three metrics. All conversions target LAS \citep{ASPRS2019} format so output format itself does not affect the comparison across vendors.

\textit{Points converted} measures how many point cloud points are successfully extracted from the raw PCAP data \citep{Harris2022} and written into the output file. Since point density varies across vendors \citep{Roriz2022}, this gives a more consistent workload measure than scan count alone.

\textit{Time taken} captures the total runtime from opening the PCAP file to completing the LAS \citep{ASPRS2019} write operation. Full PCAP datasets are several gigabytes in size, so a max\_scans parameter is introduced to process controlled portions of the capture during benchmarking rather than repeatedly converting the complete dataset.

Because vendors generate different point densities \citep{Abbasi2022}, the number of converted points varies even with the same max\_scans value. Throughput normalizes this difference by measuring points converted per unit time, allowing runtime performance to be compared across vendors under different workloads.

\subsection{Vendor Detection Evaluation}

The vendor detection module was evaluated across all eight test files described in Section~\ref{sec:experimental_setup}. Each file was passed to the pipeline without any manual vendor specification; the detection module derived the vendor solely from packet-level signals. Table~\ref{tab:detection_results} summarises the results.

\begin{table}[H]
\centering
\caption{Vendor detection results across all test files. Confidence score reflects the normalised margin between the winning vendor score and the next highest score.}
\label{tab:detection_results}
\begin{tabular}{clllcc}
\hline
\textbf{\#} & \textbf{File} & \textbf{True Vendor} & \textbf{Format} & \textbf{Detected} & \textbf{Confidence} \\
\hline
1 & hesai\_BusyRoad.pcap & Hesai & .pcap & Hesai & 48.3\% \\
2 & Outdoor\_sampledata.lvx2 & Livox & .lvx2 & Livox & 22.4\% \\
3 & StaticCarIntersection\_Livox-HAP.pcap & Livox & .pcap & Livox & 39.7\% \\
4 & OS-2-128\_...\_100425-000.pcap & Ouster & .pcap & Ouster & 39.4\% \\
5 & Urban\_Drive\_.pcap & Ouster & .pcap & Ouster & 56.9\% \\
6 & CarLoop\_Velodyne-VLP16.pcap & Velodyne & .pcap & Velodyne & 60.1\% \\
7 & Drone\_Velodyne-HDL32.pcap & Velodyne & .pcap & Velodyne & 61.5\% \\
8 & OfficeWalking\_Velodyne-VLS128.pcap & Velodyne & .pcap & Velodyne & 61.8\% \\
\midrule
\multicolumn{4}{l}{\textbf{Detection accuracy}} & \multicolumn{2}{r}{\textbf{8/8 (100\%)}} \\
\bottomrule
\end{tabular}
\end{table}

Detection was correct for all eight files across all four vendors. The three Velodyne files, spanning three different sensor models (VLP-16, HDL-32, VLS-128), all returned high confidence scores above 60\%, reflecting the distinctiveness of Velodyne's packet structure and port signature. Ouster files returned moderate confidence scores, consistent with the known ambiguity in Ouster's port range overlapping with general UDP traffic. The Livox \texttt{.lvx2} file returned the lowest confidence at 22.4\%, since extension-based signals carry less discriminative weight than packet-structure signals; however, detection remained correct. The Hesai file at 48.3\% reflects the port overlap with Velodyne on port 2368, resolved in practice by the combined contribution of magic byte and packet structure signals as discussed in Section~\ref{sec:vendor_detection}. No misclassification occurred across any test file.

\subsection{Results and Analysis}

Using the metrics defined above, three graphs are analysed in this section, each targeting a different aspect of conversion performance, mainly how workload size, parser architecture, and system limits affect conversion performance across vendors.

\subsubsection{Scaling Behaviour --- Points Converted vs Time Taken}

\begin{figure}[H]
\centering
\includegraphics[width=\linewidth]{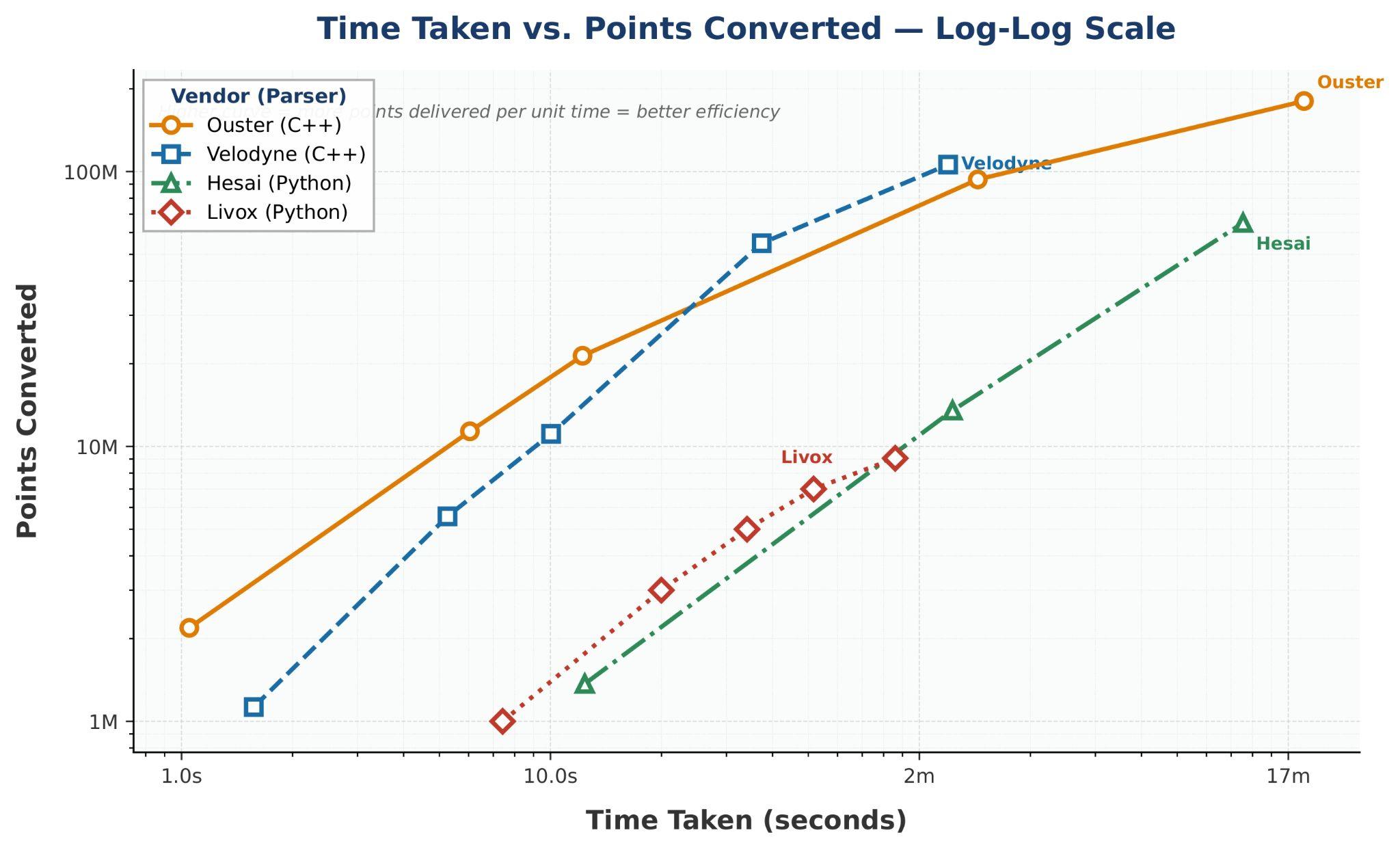}
\caption{Time taken vs. points converted (log-log scale) across all four vendors.}
\label{fig:time_vs_points}
\end{figure}

The relationship between execution time and processed point volume is shown in Figure~\ref{fig:time_vs_points} using a log-log scale, since both runtime and point counts vary significantly across vendors and datasets.

All four vendors move in the same direction --- more time leads to more points being processed. But the slope is not the same for everyone. Ouster \citep{Ouster2024SDK} and Velodyne \citep{Valgur2024} scale more efficiently at larger workloads, converting substantially more points within similar runtime windows. Hesai \citep{Hesai2025} and Livox \citep{Livox2019LVX} grow more gradually in comparison.

The Livox trend terminates earlier than the other vendors because the benchmark dataset itself is smaller in size. The available packet stream is exhausted before reaching comparable point volumes, making this a dataset limitation rather than a pipeline limitation.

Runtime increased consistently across all four vendors as workload size grew, and no abrupt slowdown or conversion instability was observed during larger runs.

\subsubsection{Average Throughput Across Vendors}

\begin{figure}[H]
\centering
\includegraphics[width=0.8\linewidth]{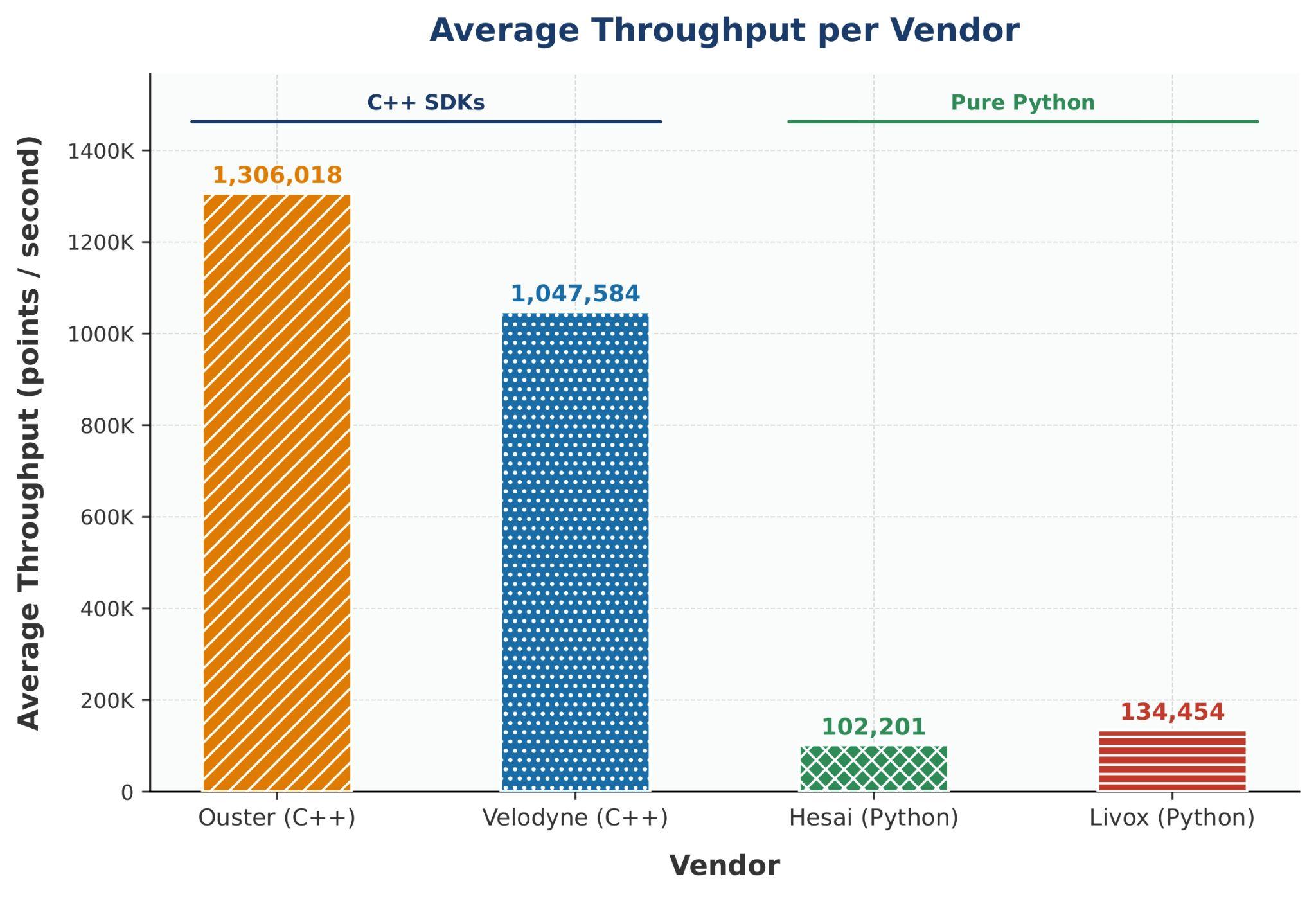}
\caption{Average throughput per vendor, showing the separation between C++ SDK-backed parsers (Ouster, Velodyne) and pure Python parsers (Hesai, Livox).}
\label{fig:avg_throughput}
\end{figure}

Figure~\ref{fig:avg_throughput} compares the average throughput observed across different max\_scans workloads for all four vendors. The max\_scans parameter controls how many sensor scans are processed during a single execution run, allowing throughput behaviour to be compared under similar workload conditions.

A clear separation appears between vendors backed by C++ SDKs \citep{Ouster2024SDK,Valgur2024} and those relying primarily on Python-based packet decoding \citep{Song2024dpkt}. Ouster \citep{Ouster2024SDK} and Velodyne \citep{Valgur2024} sustain average throughput near 1.3M and 1M points per second respectively, while Hesai \citep{Hesai2025} and Livox \citep{Livox2019LVX} remain close to 134K and 102K points per second across larger workloads.

Despite the runtime differences, conversion completeness and extracted point counts remained consistent across all four vendors during evaluation.

\subsubsection{Throughput Degradation Under Load}

\begin{figure}[H]
\centering
\includegraphics[width=\linewidth]{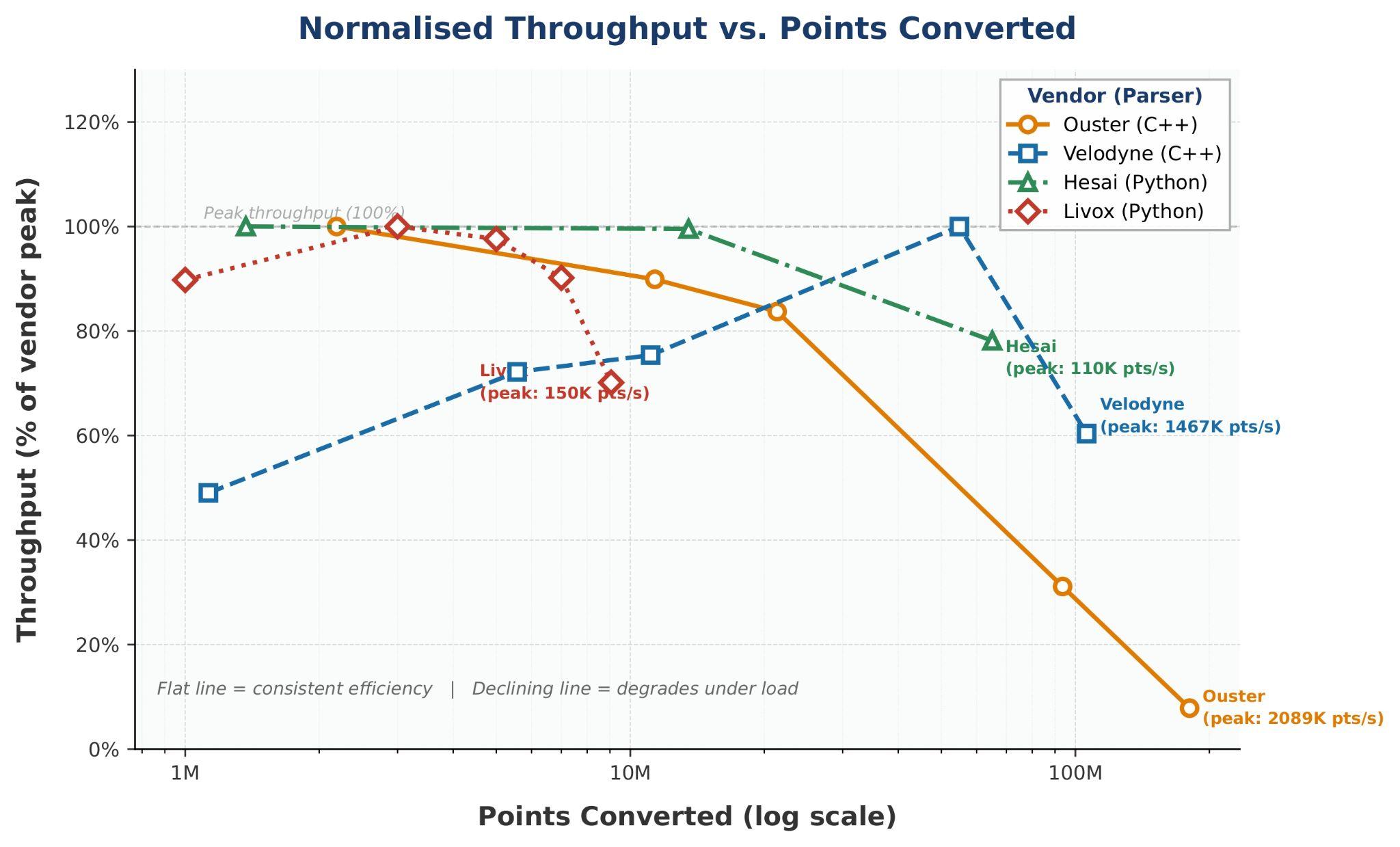}
\caption{Normalised throughput vs. points converted, expressed as a percentage of each vendor's peak throughput.}
\label{fig:normalised_throughput}
\end{figure}

Figure~\ref{fig:normalised_throughput} illustrates how throughput changes as workload size increases across different vendor pipelines. The graph highlights how parser implementation, memory usage, and output-writing overhead begin affecting sustained conversion performance at larger point counts \citep{Wang2025}.

The degradation behaviour differs noticeably across vendors. Ouster \citep{Ouster2024SDK} and Hesai \citep{Hesai2025} show a gradual throughput decline as workload size increases, mainly due to growing memory pressure and the higher cost of writing larger LAS \citep{ASPRS2019} outputs. Livox \citep{Livox2019LVX} follows a different trend: throughput initially improves as startup overhead becomes less dominant over larger runs, then begins dropping once resource pressure increases. Velodyne \citep{Valgur2024} continues scaling efficiently through medium workloads before declining at higher point counts, where memory handling and packet buffering begin limiting sustained throughput.

These throughput profiles indicate that vendor parser behaviour is the primary performance constraint rather than the conversion layer itself \citep{Lin2017}. Different vendors stress different parts of the pipeline as workload size grows, which means optimization strategies do not remain uniform across all parser backends.

\section{Threats to Validity}
\label{sec:threats}

\textbf{Detection generalisation.} Signal weights were determined empirically over the eight-file test set rather than through formal optimisation. Detection was correct across all files, but the weights may not generalise to edge cases outside the test set, such as truncated captures, non-standard firmware configurations, or vendors whose packet layouts deviate significantly from published documentation. The scoring mechanism is intentionally conservative, and the confidence score provides an explicit signal when a detection is marginal.

\textbf{Dataset scope.} Only one Hesai file was available for evaluation, reflecting the limited open distribution of raw Hesai PCAP data rather than a pipeline constraint. Conclusions about detection robustness for Hesai are accordingly limited to the single file tested. Throughput benchmarks were conducted on one hardware configuration (Intel i3, 8~GB RAM, Windows~11); performance on Linux workstations or multi-core server hardware may differ, particularly for the C++-backed vendors where OS-level scheduling behaviour affects sustained throughput.

\textbf{Throughput measurement scope.} Throughput was measured under a fixed \texttt{max\_scans} limit to control for file size differences across vendors. This does not fully characterise behaviour under sustained memory pressure for very large captures. The 8--10$\times$ gap between C++-backed and Python-based parsers is an architectural observation tied to SDK availability rather than a tuned result; both categories have room for optimisation beyond the scope of this paper.

\textbf{SDK dependency.} Correctness and performance for Ouster and Velodyne depend on the stability of the \texttt{ouster-sdk} and \texttt{velodyne-decoder} libraries. Future SDK updates or firmware-driven packet format changes may require updates to the relevant wrappers. The modular architecture is designed to contain such changes within individual wrapper implementations without affecting the core pipeline.

\section{Conclusion}
\label{conclusion}

Raw PCAP data \citep{Harris2022} from heterogeneous LiDAR deployments lacks a unified, vendor-agnostic processing path. Each manufacturer imposes distinct SDK requirements \citep{Ouster2024SDK,Valgur2024}, unique packet parsing assumptions \citep{Lin2017}, and in the case of Hesai and Livox, no production-grade open-source Python SDK for file conversion \citep{Hesai2025,Livox2019LVX}. The proposed pipeline addresses this directly: it identifies the vendor from the raw packet stream through a weighted multi-signal scoring mechanism, then routes the data through the appropriate parsing backend and writes the output to the requested standard format, without any manual vendor selection.

The pipeline was tested against real-world captures from all four vendors. Vendor detection succeeded on every dataset without manual configuration, and no parser failures were recorded across any of the evaluation runs. Performance benchmarks confirmed a clear architectural gap: C++-backed vendors (Ouster, Velodyne) reached approximately 1.3M and 1M points per second, while Python-based vendors (Hesai, Livox) remained near 134K and 102K points per second --- an 8--10$\times$ difference tied to SDK availability rather than pipeline design \citep{Lin2017}.

The pipeline still faces a few practical limitations at larger workloads. During higher point-count processing, especially with Ouster \citep{Ouster2024SDK} and Hesai \citep{Hesai2025} datasets, memory pressure begins affecting sustained throughput and total runtime \citep{Wang2025}. Additional optimisation around chunked memory management \citep{laspy2024Compression} could improve scalability for larger captures. The current wrapper structure also leaves room for adding more vendors later, since new parsers only need to integrate with the existing base interface already used throughout the pipeline. Ouster has also begun distributing sensor data in their newer proprietary OSF (Ouster Sensor Format) in place of raw PCAP, and extending the detection and ingestion logic to support OSF is a natural next step for maintaining compatibility with Ouster's current data ecosystem. Near-term improvements also include packaging the project as a PyPI \citep{PyPA2024} library so setup becomes less dependent on manual installation steps. The full implementation, including all vendor wrappers, the detection module, and evaluation scripts, is available at \citep{PatelGitHub2024}.

\section*{CRediT authorship contribution statement}
\textbf{Param Patel:} Conceptualization, Methodology, Software, Validation, Investigation, Data curation, Writing -- original draft, Visualization. \textbf{Jay Dave:} Supervision, Writing -- review \& editing. \textbf{Pratyush Chakraborty:} Supervision, Writing -- review \& editing, Project administration.

\section*{Acknowledgements}
During the preparation of this manuscript, the authors used \textbf{Claude (Anthropic)} and \textbf{ChatGPT (OpenAI)} as AI-assisted writing tools to support language refinement, structural editing, and clarity improvements of the text. These tools were used strictly as writing aids and did not contribute to the research design, methodology development, implementation, experiments, or interpretation of results. All technical content, system design, algorithms, implementation details, and experimental evaluations were developed independently by the authors. The authors reviewed, verified, and take full responsibility for the integrity, correctness, and originality of all submitted content.

\section*{Conflict of Interest}
The authors declare that they have no known competing financial interests or personal relationships that could have appeared to influence the work reported in this paper.

\bibliographystyle{plainnat}
\bibliography{sample}

\end{document}